%% file: anonymous-submission-latex-2026.tex
\documentclass[letterpaper]{article} 
\usepackage[draft]{aaai2026}  
\usepackage{times}  
\usepackage{helvet}  
\usepackage{courier}  
\usepackage[hyphens]{url}  
\usepackage{graphicx} 
\usepackage{natbib}  
\usepackage{caption} 
\usepackage{algorithm}
\usepackage{algpseudocode}
\usepackage{inconsolata}

\usepackage[utf8]{inputenc} 
\usepackage{url}            
\usepackage{booktabs}       
\usepackage{amsfonts}       
\usepackage{nicefrac}       
\usepackage{microtype}      
\usepackage{xcolor}         
\usepackage{amsmath}        
\usepackage{subcaption}
\usepackage{algpseudocode}
\usepackage{multirow}
\usepackage{tikz}
\usepackage{bbding}
\usepackage{array}
\usepackage{makecell}
\usepackage{tabularx}
\usepackage{svg}
\definecolor{deepgreen}{rgb}{0.2, 0.5, 0.2}
\definecolor{niceblue}{rgb}{0.2, 0.4, 0.8}
\definecolor{nicered}{rgb}{0.8, 0.2, 0.2}
\usepackage{tcolorbox}
\usepackage{fontawesome5}

\usepackage{amssymb}
\usepackage{dsfont}
\usepackage{newfloat}
\usepackage{listings}
\DeclareCaptionStyle{ruled}{labelfont=normalfont,labelsep=colon,strut=off} 
\floatstyle{ruled}
\newfloat{listing}{tb}{lst}{}
\floatname{listing}{Listing}
\title{GlitchMiner: Mining Glitch Tokens in Large Language Models via Gradient-based Discrete Optimization}
\author{
    Zihui Wu\textsuperscript{\rm 1},
    Haichang Gao\textsuperscript{\rm 1}\protect\thanks{Corresponding author.},
    Ping Wang\textsuperscript{\rm 1},
    Shudong Zhang\textsuperscript{\rm 1},
    Zhaoxiang Liu\textsuperscript{\rm 2, \rm 3},
    Shiguo Lian\textsuperscript{\rm 2, \rm 3}\protect\thanks{Corresponding author.}
}
\affiliations {
    \textsuperscript{\rm 1}School of Computer Science and Technology, Xidian University\\
    \textsuperscript{\rm 2}Data Science \& Artificial Intelligence Research Institute, China Unicom\\
    \textsuperscript{\rm 3}Unicom Data Intelligence, China Unicom\\
    zihui@stu.xidian.edu.cn, hchgao@xidian.edu.cn, liansg@chinaunicom.cn
}

\usepackage{bibentry}

\begin{document}

\maketitle

\begin{abstract}
Glitch tokens—inputs that trigger unpredictable or anomalous behavior in Large Language Models (LLMs)—pose significant challenges to model reliability and safety. Existing detection methods primarily rely on heuristic embedding patterns or statistical anomalies within internal representations, limiting their generalizability across different model architectures and potentially missing anomalies that deviate from observed patterns.
We introduce \textbf{GlitchMiner}, an behavior-driven framework designed to identify glitch tokens by maximizing predictive entropy. Leveraging a gradient-guided local search strategy, GlitchMiner efficiently explores the discrete token space without relying on model-specific heuristics or large-batch sampling.
Extensive experiments across ten LLMs from five major model families demonstrate that GlitchMiner consistently outperforms existing approaches in detection accuracy and query efficiency, providing a generalizable and scalable solution for effective glitch token discovery. Code is available at \url{https://github.com/wooozihui/GlitchMiner}.
\end{abstract}

\section{Introduction}
\label{sec:1}

Large Language Models (LLMs) have demonstrated remarkable capabilities and are increasingly deployed in high-stakes domains such as code generation~\cite{LLM_code1,LLM_code2,LLM_code3}, healthcare~\cite{medical1,medical2}, and education~\cite{education1,education2}. As reliance on LLMs grows, ensuring their safety and output reliability becomes imperative.

A particularly concerning threat arises from a class of anomalous inputs known as \emph{glitch tokens}~\cite{solidgoldmagikarp}. These tokens can induce LLMs to generate unexpected, erratic, or even policy-violating outputs~\cite{geiping2024coercing}, undermining trust in their behavior. For example, as illustrated in Figure~\ref{fig:glitch_example}, when prompted to repeat the word ``Mediabestanden,'' Llama2-7b-chat outputs ``hello world''—a semantically unrelated response. This phenomenon indicates that certain tokens can trigger unpredictable model behavior, even under simple and deterministic instructions.

In response, several studies have been proposed to detect glitch tokens~\cite{Magikarp, GlitchHunter, GlitchProber}. These methods typically rely on heuristic observations or statistical patterns in token embeddings or latent representations, such as small embedding norms or localized embedding clusters. Although effective in certain scenarios, such heuristics have two main limitations. First, the reliance on specific observed statistical features may not generalize well across different architectures or training setups. Second, tokens that do not conform to these empirically observed patterns may remain undetected, potentially limiting the overall coverage and robustness of these detection methods.

\begin{figure}[t] 
\centering 

\begin{tcolorbox}[
    width=\linewidth, 
    colback=blue!5,
    colframe=blue!75!black,
    title=\faUser\space Human,
    fonttitle=\bfseries
]
Please repeat the string: Mediabestanden
\end{tcolorbox}

\begin{tcolorbox}[
    width=\linewidth, 
    colback=red!5,
    colframe=red!75!black,
    title=\faRobot\space LLM,
    fonttitle=\bfseries
]
String: "hello world"
\end{tcolorbox}
\caption{An illustrative example of how glitch token causing Llama2-7b-chat to fail a simple repetition task. More examples can be found in Appendix B.}
\label{fig:glitch_example} 
\end{figure}

To address these limitations, we propose a behavior-driven perspective. Rather than searching for fixed patterns within the model, we hypothesize that glitch tokens are outliers in the model’s learned distribution, and thus elicit \emph{uncertain} predictions. This uncertainty can be quantified via the entropy of the model’s next-token distribution—high entropy indicates that the model is indecisive, providing a proxy for behavioral instability.

Building on this insight, we present \textbf{GlitchMiner}, a novel framework that identifies glitch tokens by maximizing output entropy. The core of GlitchMiner is a \emph{gradient-guided local search}, which efficiently navigates the discrete token space to identify high-entropy candidates. Unlike prior gradient-based discrete optimization strategies that suffer from inaccurate approximations or require large batch sampling~\cite{autoprompt,GCG}, our method restricts updates to neighboring tokens in the embedding space. This ensures accurate Taylor-based entropy estimates and improves search efficiency.

\paragraph{Our Contributions} are summarized as follows:
\begin{itemize}
    \item We propose a behavior-driven framework for glitch token detection, leveraging output entropy as an architecture-agnostic measure of model uncertainty to ensure broad applicability across diverse models.
    
    \item We introduce a gradient-guided local search algorithm that accurately estimates entropy gradients within the discrete token space, eliminating the dependence on large-batch sampling and improving search efficiency.
    
    \item Extensive experiments across 10 LLMs from 5 major model families demonstrate that GlitchMiner consistently surpasses state-of-the-art methods in detection accuracy, establishing it as a robust and effective solution for glitch token detection.
\end{itemize}

The remainder of the paper is organized as follows: Section~\ref{sec:2} reviews related work; Section~\ref{sec:3} describes the GlitchMiner methodology; Section~\ref{sec:4} presents experiments and ablations; and Section~\ref{sec:5} concludes with future directions.

\section{Background and Related Work}
\label{sec:2}

\subsection{Glitch Token Definition}

In practice, the identification of glitch tokens relies on specific behavioral tests. The most common approach, used in prior work~\cite{Magikarp, GlitchHunter, GlitchProber}, is the \textbf{repetition task}. In such a task, a model is prompted to repeat a given token $t$; a failure to accurately reproduce the token is taken as evidence of it being a glitch. This single-test method, however, is fragile. The outcome can be highly sensitive to the specific wording of the prompt, leading to inconsistent results and potential false positives.

To establish a more robust and replicable definition, we move beyond single-prompt evaluations. We classify a token $t$ as a \emph{glitch token} only if it fails the repetition task consistently across a diverse set of $m$ prompt templates. This stricter criterion is formalized as follows: given a set of $m$ templates, $h_i(\cdot)\}_{i=1}^m$, the failure rate for a token $t$ is defined as:

\begin{equation}
\label{eq:fail_rate}
\text{fail}(t) := \frac{1}{m} \sum_{i=1}^{m} \mathds{1}\left[ \arg\max_{v \in \mathcal{V}} \, P(v \mid h_i(t)) \neq t \right]
\end{equation}

A token is then identified as a glitch if and only if its failure rate is absolute, i.e., $\text{fail}(t) = 1$. This rigorous cross-verification process effectively filters out template-specific artifacts, ensuring that our analysis focuses exclusively on consistently unstable tokens.

\subsection{Glitch Token Detection}

A series of methods have been proposed to \emph{efficiently rank or localise} glitch tokens in large vocabularies:

\textbf{Magikarp}~\cite{Magikarp} adopts a lightweight heuristic, screening tokens with atypically small $\ell_2$ embedding norms before verifying them via the repetition task.

\textbf{GlitchHunter}~\cite{GlitchHunter} observes that glitch tokens tend to cluster in embedding space. It builds a token-embedding graph and applies Leiden clustering~\cite{traag2019louvain}, followed by iterative hypothesis testing to refine each cluster.

\textbf{GlitchProber}~\cite{GlitchProber} shifts the focus from embeddings to \emph{internal activations}. It projects hidden states and attention outputs with PCA, then trains SVM classifiers to flag activation outliers; a mitigation step masks offending neurons at inference time.

While effective, these approaches share a common limitation: they rely heavily on heuristic observations of embedding or activation patterns. Such reliance may reduce their generalizability across different architectures, and tokens not exhibiting these typical patterns may remain undetected.

In contrast, GlitchMiner employs gradient-guided discrete optimization to maximize output entropy, an architecture-agnostic indicator of predictive uncertainty that facilitates broad generalization across diverse LLMs.

\subsection{Gradient-based Discrete Optimization}
Gradient-based discrete optimization methods \cite{hotflip, autoprompt, GCG, PEZ} leverage gradient information to predict how individual tokens impact the loss function. These approaches typically treat the one-hot encoding of tokens or token embeddings as continuous vectors to compute gradients, guiding token replacements for optimization.

\textbf{HotFlip} \cite{hotflip} uses the one-hot encoding of the tokens to calculate the gradients and selects the token with the largest negative gradient to replace the current token, with the goal of minimizing the loss. However, it only evaluates one candidate token per iteration, which can lead to suboptimal predictions and reduced accuracy.

\textbf{AutoPrompt} \cite{autoprompt} improves upon HotFlip by evaluating multiple candidate tokens in each iteration. Instead of relying on gradients from one-hot encodings, it utilizes token embedding gradients for loss estimation, enhancing prediction accuracy by considering a broader range of potential token replacements.

\textbf{GCG} \cite{GCG} extends HotFlip by incorporating multi-candidate token selection, similar to AutoPrompt, but it still uses the one-hot encoding of tokens to compute gradients for loss estimation. Notably, GCG has been applied to automated \emph{jailbreaks} \cite{shen2023anything} in LLMs, efficiently searching for adversarial suffixes.

AutoPrompt and GCG both rely on \textbf{large batch sampling} to mitigate inaccuracies in gradient-based prediction. We identified that these inaccuracies arise from the inaccuracy of Taylor expansions when input tokens are distant from the original points. This overlooks a fundamental condition of Taylor approximation: its accuracy is highest for points close to the reference point.

Building on these works, we introduce a \textbf{local search strategy} in our approach. This improvement enables us to achieve high precision in gradient estimation without relying on large batch sampling, by focusing on a smaller, localized token space. By addressing the core issue of Taylor approximation accuracy, our method allows for more efficient and accurate exploration of the token space, which is particularly valuable for glitch token detection.

\section{Methodology}
\label{sec:3}

Our goal is to develop a method that can automatically and efficiently discover glitch tokens in any given LLM. To this end, we propose \textbf{GlitchMiner}, a novel framework that formulates glitch token discovery as a behavior-driven optimization problem.

GlitchMiner systematically mines for these tokens through a powerful iterative loop, outlined in Algorithm~\ref{alg:framework}. Each iteration first performs \textbf{Gradient-Guided Selection} to pinpoint tokens with the highest predicted output entropy. These candidates then immediately undergo \textbf{Robust Verification} to determine if they are genuine glitch tokens.

\begin{algorithm}[t]
    \caption{The GlitchMiner Pipeline}
    \label{alg:framework}
    \begin{algorithmic}[1]
    \State \textbf{Input:} LLM model $f(\cdot)$, Full vocabulary $\mathcal{T}$, Max iterations $N$
    \State \textbf{Output:} Verified glitch token set $\mathcal{G}$
    
    \State \textcolor{gray}{\# Initialization}
    \State $\mathcal{T}^* \gets \text{PreFilter}(\mathcal{T})$ \Comment{Exclude irrelevant tokens, see Sec. 3.3}
    \State $\mathcal{G} \gets \emptyset$
    
    \State \textcolor{gray}{\# Iterative Mining Loop}
    \For{$i$ from $1$ to $N$}
        \State \textcolor{gray}{\# 1. Gradient-guided selection of promising candidates}
        \State $\mathcal{T}_c \gets \mathcal{T} \setminus (\mathcal{T}^* \cup \mathcal{G})$ \Comment{Define current candidate set}
        \State $\mathcal{B} \gets \text{SelectCandidateBatch}(\mathcal{T}_c)$ \Comment{See Sec. 3.1}
        
        \State \textcolor{gray}{\# 2. Robust verification of each candidate}
        \For{each token $t \in \mathcal{B}$}
            \If {VerifyIsGlitch($t$)} \Comment{See Sec. 3.2}
                \State $\mathcal{G} \gets \mathcal{G} \cup \{t\}$
            \EndIf
            \State $\mathcal{T}^* \gets \mathcal{T}^* \cup \{t\}$ \Comment{Mark token as processed}
        \EndFor
    \EndFor
    \State \textbf{return} $\mathcal{G}$
    \end{algorithmic}
\end{algorithm}

\subsection{Gradient-Guided Candidate Selection}
\label{sec:candidate_selection}

The core of GlitchMiner's discovery engine is the candidate selection process. To ensure the overall mining is efficient, this selection is not random; instead, it employs a local search strategy to intelligently find tokens that maximize model confusion, which we measure using output entropy.

\textbf{Optimization Objective}.
To formally quantify a token's impact on model uncertainty, we calculate its output entropy within a repetition task. We use the following template for this optimization process:

\begin{quote}
User: Please repeat the string: ``\{token\}"\\
Assistant: Sure, the string is: ``\{token\}
\end{quote}

\begin{figure}[t]
 \centering
 \includegraphics[width=1\linewidth]{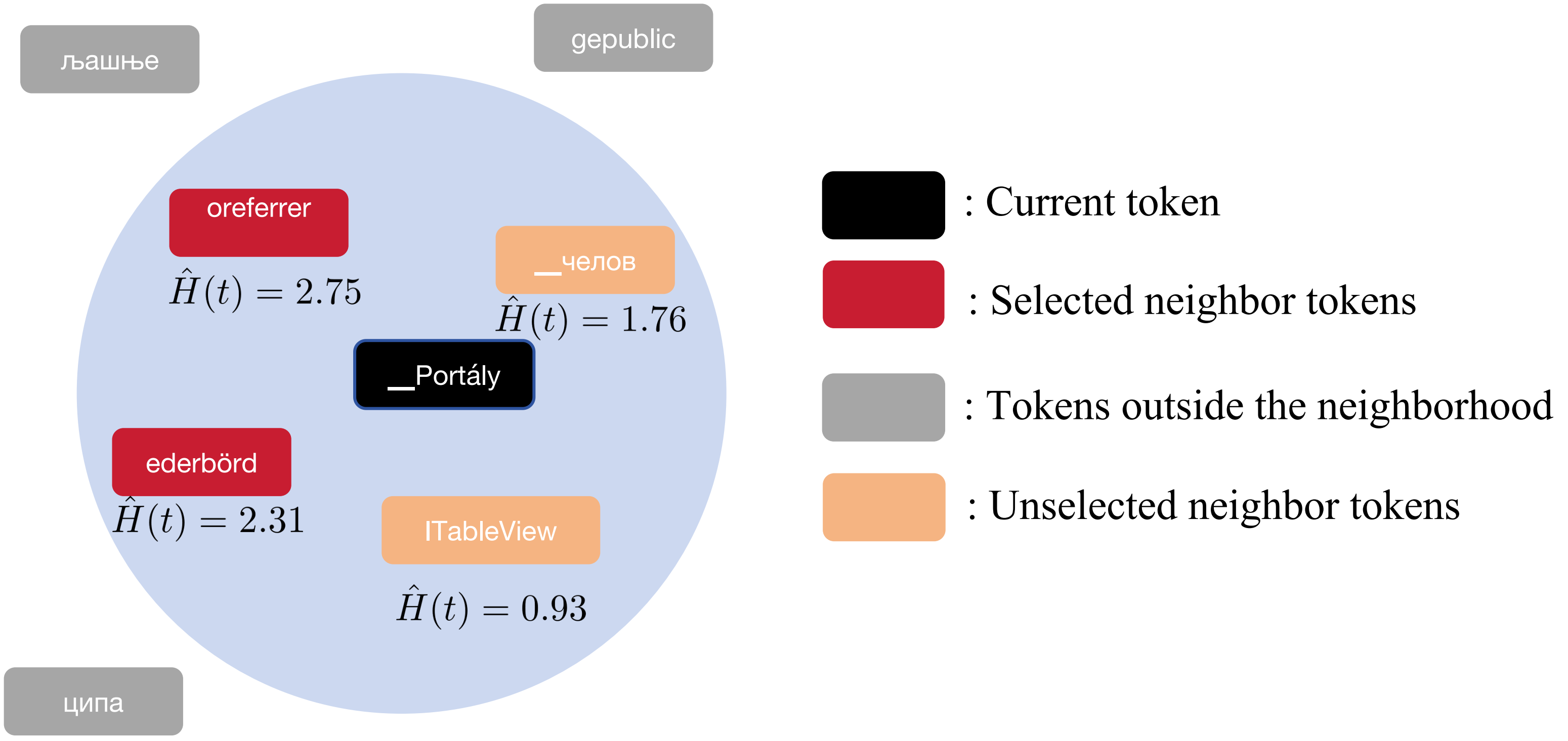} 
 \caption{Visualization of GlitchMiner’s local search process. The \textbf{pivot token} (black, $t_c$) serves as the reference point. Its \textbf{neighbor tokens} (orange and red) represent the $K=4$ closest tokens in embedding space. Among these, the \textbf{candidate batch tokens} (red) are the top $B=2$ tokens with the highest \textbf{approximate entropy values}, estimated via first-order Taylor approximation. Tokens outside the neighborhood (gray) are excluded to maintain approximation accuracy and computational efficiency.}
 \label{fig:local_search}
\end{figure}

Here, the first \{token\} represents the input token $t$ being evaluated,
and the second \{token\} is the next predicted token by LLM. To ensure that the input token appears directly as the model's next prediction, we prefill the assistant's response with the phrase \emph{Sure, the string is: ``}.
The entropy is computed based on the model’s next-token distribution $p(v \mid \mathbf{h}(t))$, where $\mathbf{h}(t)$ denotes the prompt generated by inserting $t$ into the repetition template $h$. The entropy is thus:
\[
H(t) = -\sum_{v \in \mathcal{V}} p(v \mid \mathbf{h}(t)) \log p(v \mid \mathbf{h}(t))
\]

Our goal is to find a batch of tokens $\mathcal{B}$ from the current set of available tokens $\mathcal{T}_c$ that maximizes the total entropy:
\[
\mathcal{B} = \arg\max_{\mathcal{B} \subset \mathcal{T}_c, |\mathcal{B}| = B} \sum_{t \in \mathcal{B}} H(t)
\]
where $\mathcal{T}_c = \mathcal{T} \setminus (\mathcal{T}^* \cup \mathcal{G})$, with $\mathcal{T}^*$ being the set of filtered (see Section 3.3) or previously verified non-glitch tokens, and $\mathcal{G}$ being the set of glitch tokens found.

\textbf{Local Search Strategy}.
To efficiently solve this optimization problem, we introduce a \textbf{local search strategy} that addresses the limitations of global Taylor approximations. This process begins with an initial pivot token $t_c$ and iteratively refines the search. In each step:

\begin{enumerate}
    \item A local neighborhood $\mathcal{N}_K(t_c)$ is defined, consisting of the nearest neighbors $K$ to the current pivot token $t_c$ in the embedding space.
    \item Within this neighborhood, the entropy of each candidate token $t \in \mathcal{N}_K(t_c)$ is estimated using a first-order Taylor approximation:
    \[
    \hat{H}(t) \approx H(t_c) + \nabla_e H(t_c)^\top (e_t - e_{t_c})
    \]
    where $e_t$ and $e_{t_c}$ are the respective embedding vectors. This approximation avoids the high cost of exact entropy calculations for all neighbors.
    \item A batch $\mathcal{B}$ of $B$ tokens with the highest approximated entropy $\hat{H}(t)$ is selected from the neighborhood. This batch is returned for verification.
    \item To guide the next search step, the \textit{actual} entropy $H_t$ is computed for the tokens in $\mathcal{B}$, and the one with the highest actual entropy becomes the new pivot, $t_c$.
\end{enumerate}

Figure~\ref{fig:local_search} visualizes this process.
This local search strategy significantly improves the accuracy of entropy estimation by focusing on tokens close to the pivot token. This entire process corresponds to the `SelectCandidateBatch' function in Algorithm~\ref{alg:framework}.

\begin{table*}[t]
    \centering
    \begin{tabularx}{\textwidth}{lX}
        \toprule
        \textbf{Model Family} & \textbf{Model Names} \\
        \midrule
        Llama Models   & Llama-3.1-8B-Instruct , Llama-2-7B-chat-hf \\
        Qwen Models   & Qwen2.5-7B-Instruct , Qwen2-7B-Instruct  \\
        Gemma Models  & Gemma-2-2b-it, Gemma-2-9b-it \\
        Phi-3 Models  & Phi-3-mini-128k-instruct, Phi-3.5-mini-instruct \\
        Mistral Models & Mistral-7B-Instruct-v0.3, Mistral-Nemo-Instruct-2407 \\
        \bottomrule
    \end{tabularx}
    \caption{Test LLMs used in the experiments.}
    \label{tab:target_llms}
\end{table*}
\begin{table*}[t]
    \centering
    \scriptsize 
    \begin{tabular*}{\textwidth}{@{\extracolsep{\fill}} l|c|c|c|c}
        \toprule
        \textbf{Model} & \textbf{Metric} & \textbf{GlitchHunter} & \textbf{Magikarp} & \textbf{GlitchMiner (ours)} \\
        \midrule
        \multirow{2}{*}{Llama-3.1-8B-Instruct} & Detected@1000 & 25 & \textbf{664} & 568 \\
        & Detected@2000 & 56 & 935 & \textbf{1164} \\
        \hline
        \multirow{2}{*}{Llama-2-7B-chat-hf} & Detected@1000 & 61 & 100 & \textbf{319} \\
        & Detected@2000 & 126 & 186 & \textbf{532} \\
        \hline
        \multirow{2}{*}{Qwen2.5-7B-Instruct} & Detected@1000 & 75 & 1000 & \textbf{1000} \\
        & Detected@2000 & 180 & \textbf{1893} & 1839 \\
        \hline
        \multirow{2}{*}{Qwen2-7B-Instruct} & Detected@1000 & 96 & 999 & \textbf{1000} \\
        & Detected@2000 & 191 & 1842 & \textbf{1847} \\
        \hline
        \multirow{2}{*}{Gemma-2-2b-it} & Detected@1000 & 23 & 678 & \textbf{744} \\
        & Detected@2000 & 35 & 984 & \textbf{1019} \\
        \hline
        \multirow{2}{*}{Gemma-2-9b-it} & Detected@1000 & 29 & 623 & \textbf{775} \\
        & Detected@2000 & 45 & 983 & \textbf{1089} \\
        \hline
        \multirow{2}{*}{Phi-3.5-mini-instruct} & Detected@1000 & 20 & 393 & \textbf{396} \\
        & Detected@2000 & 44 & 496 & \textbf{516} \\
        \hline
        \multirow{2}{*}{Phi-3-mini-128k-instruct} & Detected@1000 & 26 & 398 & \textbf{404} \\
        & Detected@2000 & 55 & 489 & \textbf{517} \\
        \hline
        \multirow{2}{*}{Mistral-7B-Instruct-v0.3} & Detected@1000 & 6 & 110 & \textbf{219} \\
        & Detected@2000 & 19 & 130 & \textbf{302} \\
        \hline
        \multirow{2}{*}{Mistral-Nemo-Instruct-2407} & Detected@1000 & 48 & 574 & \textbf{695} \\
        & Detected@2000 & 79 & 918 & \textbf{976} \\
        \midrule
        \textbf{Average} & Detected@1000 & 40.9 & 553.9 & \textbf{612.0} \\
        & Detected@2000 & 83.0 & 885.6 & \textbf{980.1} \\
        \bottomrule
    \end{tabular*}
    \caption{Detected@1000 and Detected@2000 comparison of methods across different models.}
    \label{tab:detected_results}
\end{table*}

\subsection{Robust Verification of Candidates}
\label{sec:verification}

Once a batch of promising candidates is selected, each one undergoes a rigorous verification step (`VerifyIsGlitch' in our pipeline) to filter out false positives. A candidate is confirmed as a true glitch only if it fails the repetition task across a set of diverse prompt templates. To ensure this robustness, our verification process uses $m=3$ templates for cross-validation: in addition to the template used for optimization (Sec. 3.1), we employ two others adapted from prior work~\cite{Magikarp,GlitchHunter}. A token is formally classified as a glitch if and only if its failure rate (as defined in Equation~\ref{eq:fail_rate}) is 1. The additional prompt templates and analysis of false positives are provided in Appendix A.1 and Appendix A.2, respectively.

\begin{figure*}[t]
    \centering
    \includegraphics[width=1\linewidth]{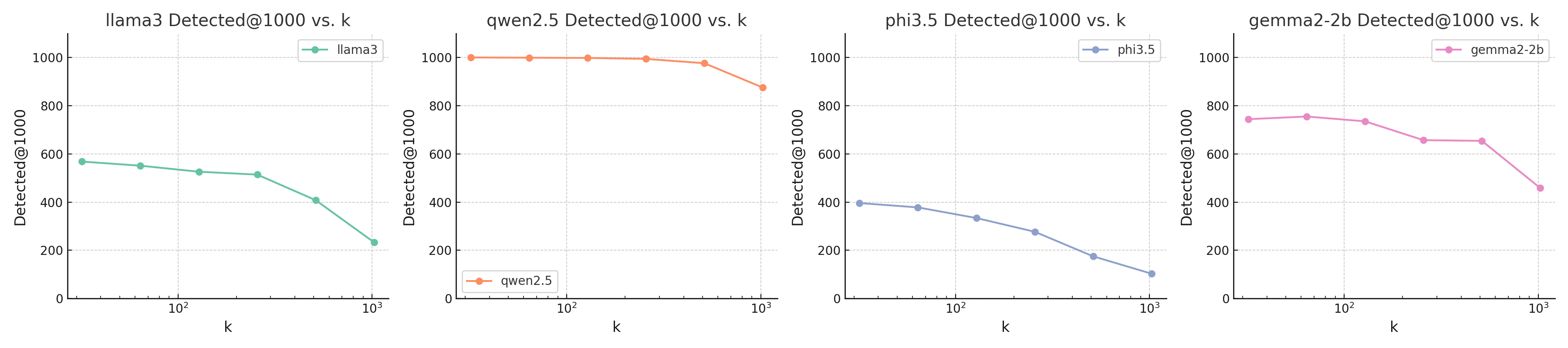}
    \caption{Impact of different Neighborhood Size $K$ on GlitchMiner's performance}
    \label{fig:k_ablation}
\end{figure*}
\begin{figure*}[t]
    \centering
    \includegraphics[width=1\linewidth]{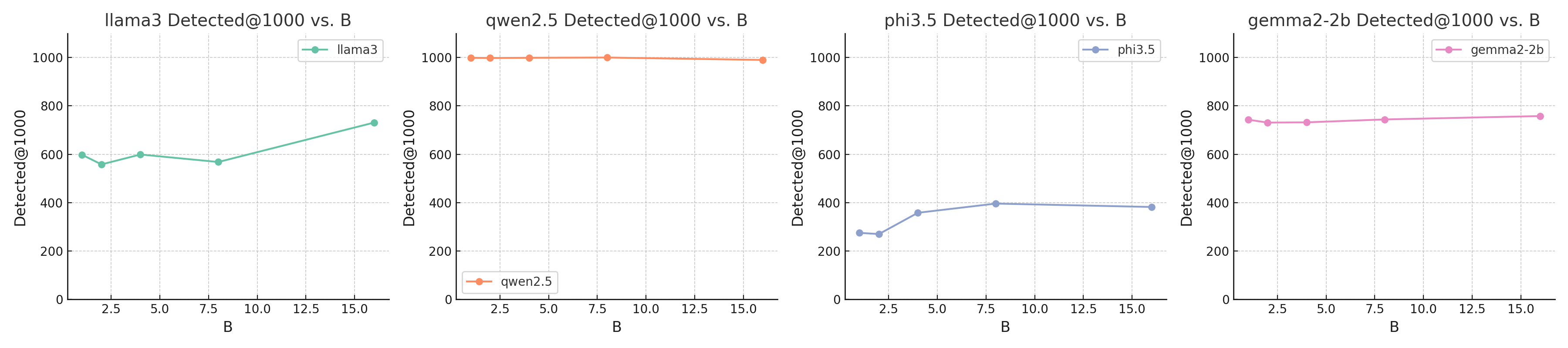}
    \caption{Impact of different Batch Size $B$ on GlitchMiner's performance}
    \label{fig:b_ablation}
\end{figure*}

\subsection{Practical Implementation: Search Space Pre-filtering}
\label{sec:filtering}

To maximize search efficiency, following \cite{land2024fishing}, we perform a one-time pre-filtering of the vocabulary (`PreFilter' in Algorithm~\ref{alg:framework}). This step removes tokens that pose no real-world risk, allowing GlitchMiner's overall search to focus its resources. We filter three categories:
\begin{itemize}
    \item \textbf{SPECIAL Tokens}: Predefined symbols like \texttt{[BOS]} or \texttt{</s>}. These are filtered out because they serve a reserved functional role for the model (e.g., indicating the start of a sequence) rather than representing user-generated text.

    \item \textbf{UNDECODEABLE Tokens}: Tokens that correspond to byte sequences that cannot be decoded into a valid string, often because they violate encoding standards like UTF-8. They are excluded as they do not map to any meaningful user-generated text.

    \item \textbf{UNREACHABLE Tokens}: Tokens that exist in the vocabulary but can never be generated by the model's tokenizer from any text input. 
\end{itemize}
Further explanations of these token categories are provided in Appendix A.3.


\section{Experiments}
\label{sec:4}
\subsection{Experimental Setup}
\textbf{Evaluated LLMs}. We used a diverse set of LLMs from five different model families to evaluate the performance of our glitch token detection approach. The selected models include Meta's Llama series \cite{llama2, llama3_1}, Alibaba's Qwen models \cite{qwen2, qwen2_5}, Google's Gemma models \cite{gemma2}, Microsoft's Phi-3 models \cite{phi3}, and Mistral models \cite{mistral_7B, mistral2024nemo}. The details are presented in Table \ref{tab:target_llms}.

\textbf{Evaluation Metrics}. We evaluate our glitch token detection method using the \textbf{Detected@N} metric, which counts the number of true glitch tokens identified within the top $N$ predictions. For instance, Detected@1000 measures how many glitch tokens are found among the top 1000 candidates. This metric balances detection accuracy and query efficiency, reflecting a method’s practical effectiveness under fixed query budgets. Comparing Detected@N values thus provides a direct measure of each method’s ability to maximize glitch token discovery while minimizing computational resources, making it well-suited for real-world applications.
\begin{figure*}[h!]
    \centering
    \includegraphics[width=0.8\linewidth]{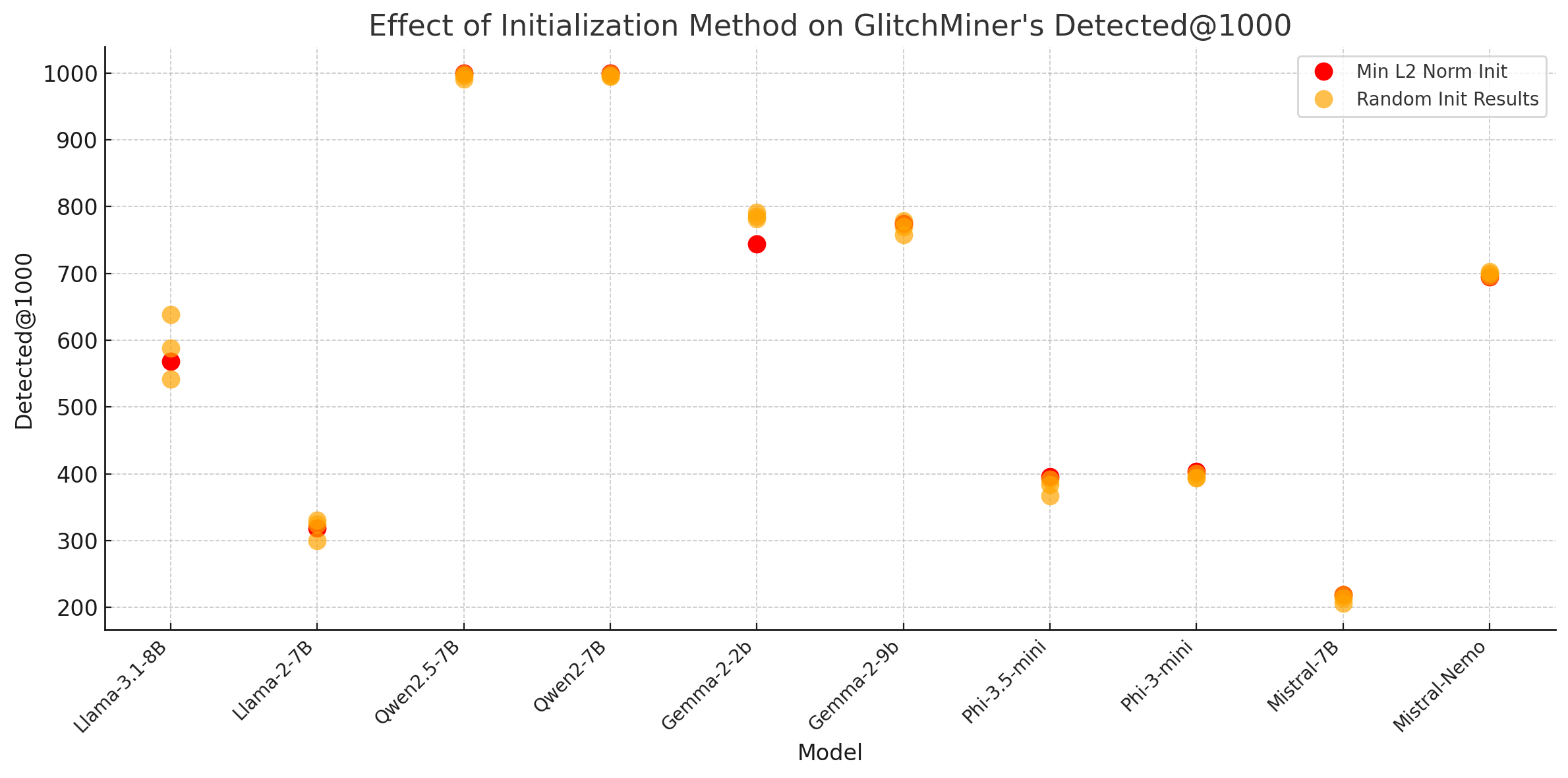}
    \caption{Effect of Initialization Method on GlitchMiner's Detected@1000 score.}
    \label{fig:init}
\end{figure*}

\begin{figure*}[h!]
    \centering
    \includegraphics[width=0.8\linewidth]{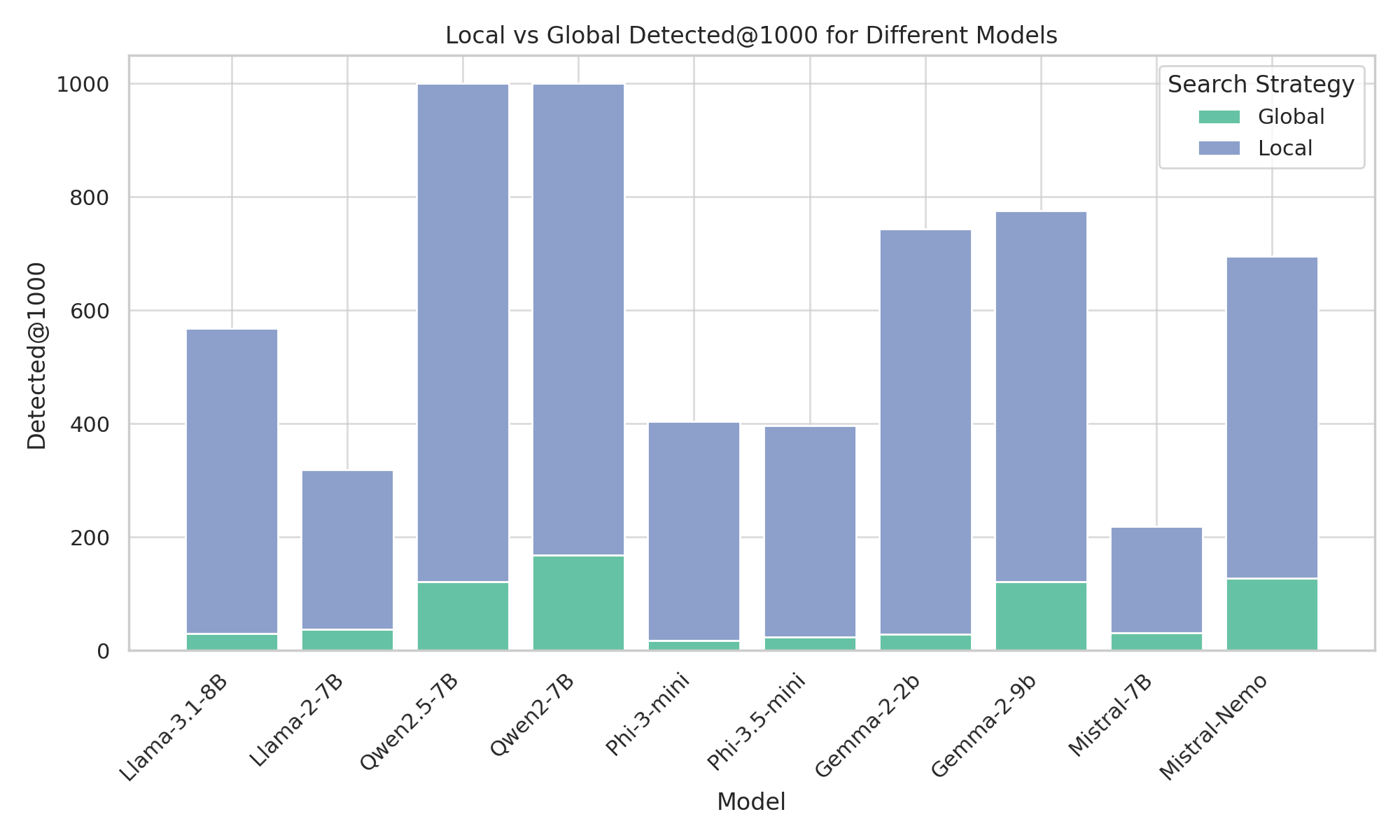}
    \caption{Comparison of GlitchMiner performance with and without local search strategy}
    \label{fig:local_vs_global}
\end{figure*}

\textbf{Baselines}. 
We compare our proposed glitch token detection method with two state-of-the-art approaches: GlitchHunter \cite{GlitchHunter} and Magikarp \cite{Magikarp}. These methods serve as the primary benchmarks for evaluating our approach.

Although GlitchProber \cite{GlitchProber} is another relevant method, it follows a fundamentally different approach by pre-collecting a subset of glitch tokens to train a classifier, introducing a supervised learning component. In contrast, GlitchMiner, along with GlitchHunter and Magikarp, uses heuristic-based methods to detect glitch tokens without relying on labeled data or additional classifier training. This methodological difference makes a direct comparison less meaningful, so we focus our evaluation on methods that align more closely with our unsupervised approach.

\textbf{Parameter Settings}. In our implementation of GlitchMiner, we use $K$=32 and $B$=8 as the default parameters. These values were chosen based on empirical testing to balance computational efficiency and detection effectiveness. Specifically, $K$=32 defines the size of the local neighborhood considered in each iteration, while $B$=8 determines the batch size for entropy computation. These settings have shown to provide a good trade-off between exploration of the token space and exploitation of local information across various model architectures.

\textbf{Initialization Strategy in Experiments}. To ensure stable and consistent comparisons across runs, we initialize the search with the token exhibiting the smallest $\ell_2$ norm in the embedding space, based on prior observations that such tokens often exhibit glitch-like behaviors. However, as shown in Figure \ref{fig:init}, we found that GlitchMiner remains robust to different initialization choices, achieving similar performance even with random starting points.

\subsection{Main Results}

The performance of GlitchMiner against the baselines is detailed in Table~\ref{tab:detected_results}. The results provide strong evidence for the effectiveness of our entropy-guided search, showing that GlitchMiner consistently discovers more glitch tokens than both GlitchHunter and Magikarp under fixed query budgets.

On average, GlitchMiner outperforms the strongest baseline, Magikarp, by 10.7\% on the Detected@2000 metric. This performance advantage is not uniform but reveals an important trend: while Magikarp's simple norm-based heuristic can be effective for an initial, low-budget scan (e.g., on Llama-3.1-8B), GlitchMiner's more sophisticated local search strategy consistently proves more fruitful as the search budget expands. For instance, on Llama-2-7B-chat-hf, GlitchMiner finds nearly three times as many glitches as Magikarp (532 vs. 186), demonstrating its superior ability to uncover less obvious candidates that simple heuristics miss. 

These findings validate that framing glitch detection as a behavior-driven optimization problem is a more robust and generalizable strategy than relying on static, model-specific patterns.

\begin{figure*}[h!]
  \centering
  \includegraphics[width=0.8\textwidth]{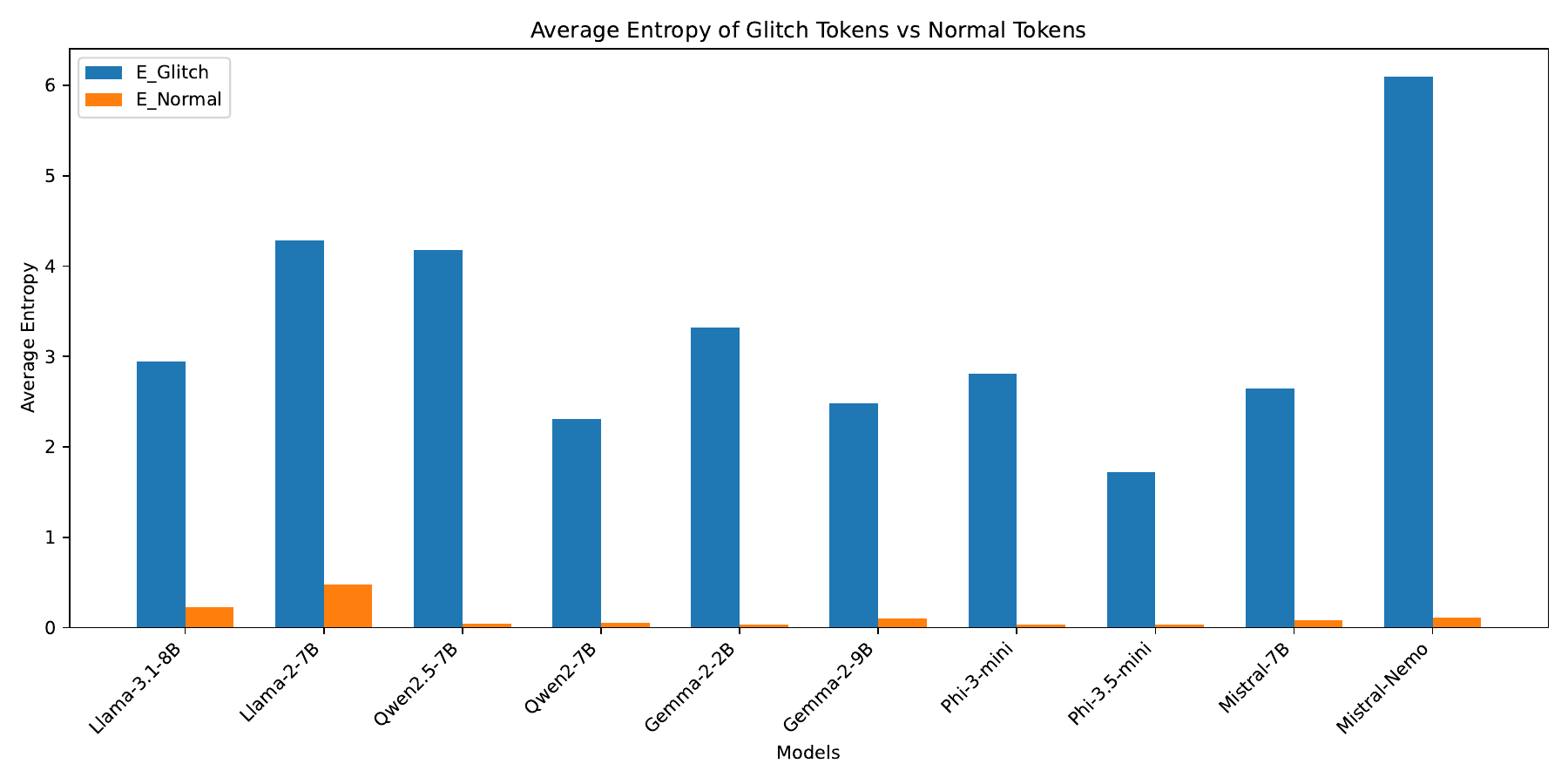}
  \caption{Average entropy comparison between glitch tokens and normal tokens across different models. Glitch tokens have higher entropy, indicating greater uncertainty in the model's predictions for these tokens.}
  \label{fig:entropy_analysis}
\end{figure*}

\subsection{Ablation Study}
To evaluate the contributions of key components in GlitchMiner, we conducted ablation studies focusing on the local search strategy, neighborhood size \(K\), batch size \(B\), and initialization token.

\textbf{Effect of Local Search}. 
The local search strategy plays a crucial role in enhancing GlitchMiner's ability to detect glitch tokens by improving the precision of the Taylor approximation. Without local search, detection accuracy drops significantly (Figure \ref{fig:local_vs_global}), as global search lacks the necessary granularity to maintain precise approximations within the token space.

\textbf{Effect of Neighborhood Size}. We analyzed the impact of neighborhood size \(K\) on detection performance. As shown in Figure \ref{fig:k_ablation}, increasing \(K\) generally leads to a decline in Detected@1000 values across models. This trend indicates that as \(K\) grows, the Taylor approximation becomes less effective, resulting in reduced prediction accuracy.

\textbf{Effect of Batch Size}. As shown in Figure \ref{fig:b_ablation}, the performance of GlitchMiner remains relatively stable as batch size \(B\) increases. Notably, even with \(B=1\), GlitchMiner achieves effective detection results, indicating that it can make accurate predictions without relying on a large batch size.

\textbf{Effect of Initialization Token}. As shown in Figure \ref{fig:init}, GlitchMiner's performance remains stable across different initialization tokens. The red dots represent the minimum $\ell_2$ norm initialization, while the orange dots show three random trials. For most models, random initialization results are close to the minimum $\ell_2$ norm, indicating that GlitchMiner achieves consistent detection accuracy regardless of the initialization approach.

\subsection{Token Entropy Analysis}

To further validate the effectiveness of our entropy-based approach in detecting glitch tokens, we conducted an entropy analysis comparing glitch tokens and normal tokens across different models. For each model, we computed the average entropy of glitch tokens (\(E_{\text{Glitch}}\)) and normal tokens (\(E_{\text{Normal}}\)).

Figure~\ref{fig:entropy_analysis} presents the comparison of average entropy values between glitch tokens and normal tokens for each evaluated model. As shown in the figure, glitch tokens consistently exhibit significantly higher entropy than normal tokens across all models.

This pronounced difference in entropy values indicates that models are more uncertain when predicting glitch tokens compared to normal tokens. The higher entropy of glitch tokens validates our hypothesis that maximizing entropy effectively guides the search towards tokens that are challenging for the model to predict.

Moreover, the consistent pattern of higher entropy for glitch tokens across diverse model families—including Llama, Qwen, Gemma, Phi-3, and Mistral—demonstrates the generality and robustness of our entropy-based approach. This suggests that our method can be effectively applied to a wide range of LLMs with different architectures and tokenization strategies.

These findings validate the effectiveness of GlitchMiner's entropy-based optimization in efficiently detecting glitch tokens by focusing on areas of high prediction uncertainty within the model.


\section{Conclusion}
\label{sec:5}
In this paper, we introduced GlitchMiner, a novel framework that detects glitch tokens in LLM by reframing the problem as a behavior-driven optimization task. Departing from prior work that relies on static patterns, our approach identifies anomalous tokens by searching for inputs that maximize the model's predictive uncertainty, measured by output entropy. We operationalize this principle with an efficient gradient-guided local search strategy that can accurately pinpoint high-entropy candidates in the vast token space. Extensive experiments on 10 diverse LLMs demonstrated that GlitchMiner significantly outperforms state-of-the-art baselines in both detection accuracy and efficiency. Our work provides a more robust and generalizable tool for auditing and enhancing the reliability of LLMs, with future work pointing towards mitigation strategies and adaptation for black-box models.

\section*{Acknowledgments}
We would like to express our sincere gratitude to Sander Land for his valuable insights and feedback
on this work, particularly regarding the evaluation metrics. We wish to thank Guanlin Li for
his valuable suggestions on our manuscript.
This work was supported in part by the National Key R\&D Program of China (2023YFB3107505), in part by Shaanxi Natural Science Funds for Distinguished Young Scholars(2023-JC-JQ-52), in part by the Natural Science Foundation of China (62302371), in part by the Postdoctoral Fellowship Program of CPSF (GZC20232035), and in part by the China Postdoctoral Science Foundation (2025M771552). 

\bibliography{aaai2026}

\newpage

\input{appendix}

\end{document}

%% file: appendix.tex
\appendix

\newpage
\begin{table*}[h]
\centering
\begin{tabularx}{\textwidth}{ll>{\ttfamily}lX}
\toprule
\textbf{Model} & \textbf{Token ID} & \textbf{Token String} & \textbf{Model's Response} \\
\midrule
Qwen2.5-7B-Instruct & 83969 & PostalCodesNL & ": " \\
\addlinespace
Qwen2.5-7B-Instruct & 88372 & useRalative & ": " \\
\addlinespace
Qwen2.5-7B-Instruct & 78323 & thuisontvangst & ": " \\
\midrule
Llama-2-7b-chat & 24291 & IABot & Of course! Here is the string "@\{":\newline\newline \&=\&=\newline\newline I hope this helps! Let me know if you have any other questions. \\
\addlinespace
Llama-2-7b-chat & 17916 & abestanden & Sure! Here is the repetition of the string: Wikimedia \\
\addlinespace
Llama-2-7b-chat & 12731 & ederbörd & Sure! The string "pon" repeats 3 times. \\
\midrule
Mistral-7B-Instruct-v0.3 & 20016 & NdEx & I'm sorry, but the string "tcx" cannot be repeated as is because it does not contain any characters or symbols that can be replicated. It may have been entered incorrectly, as it appears to consist only of backslashes and may not represent a meaningful string in its current form. Could you please check and provide the correct string if possible? \\
\addlinespace
Mistral-7B-Instruct-v0.3 & 26668 & iNdEx & I'm sorry, but the string " /******/iji\_\_[" cannot be repeated as it is not a valid or meaningful string in standard English or most programming languages. It appears to contain some uncommon characters that could represent specific instructions in certain contexts, such as within JSON or in some esoteric programming languages. Without more information about its intended usage, it's difficult to provide an accurate repetition. \\
\addlinespace
Mistral-7B-Instruct-v0.3 & 8709 & ICENSE & I'm sorry, it seems there's a typo in your request. The correct spelling is "LICENSE." Here it is again: LICENSE. \\
\bottomrule
\end{tabularx}
\caption{Examples of Glitch Tokens Discovered by GlitchMiner}
\label{tab:glitch_examples}
\end{table*}

\section{Experimental Details}
\label{sec:appendix}

\subsection{Verification Templates}
\label{sec:appendix_templates}

As stated in the main text, our robust verification process relies on three distinct templates for cross-validation to ensure reliability. A token is confirmed as a true glitch only if it fails the repetition task across all three. The first template is the one used during the optimization phase, while the other two are adapted from prior work to test for robustness under different phrasings. The structure of all three templates is detailed below for clarity and reproducibility.

\subsubsection*{Template 1: Primary Optimization Template}

\noindent\textbf{Description:} This is the primary template used to calculate output entropy during the gradient-guided search.
\begin{quote}
\textbf{User Prompt:} \texttt{Please repeat the string: "\{token\}"} \\
\textbf{Assistant Prefill:} \texttt{Sure, the string is: "` }
\end{quote}

\subsubsection*{Template 2: GlitchHunter Adaptation (Direct Repetition)}

\noindent\textbf{Description:} This template, adapted from GlitchHunter, directly asks the model to repeat the target token in a simple question-answer format.
\begin{quote}
\textbf{User Prompt:} \texttt{Question: Please can you repeat the string '\{token\}'?} \\
\textbf{Assistant Prefill:} \texttt{Of course! Here is the repeated string:`}
\end{quote}

\subsubsection*{Template 3: Magikarp Adaptation (Few-shot Consistency)}

\noindent\textbf{Description:} Adapted from Magikarp, this is a more complex few-shot template that first establishes a persona of a "helpful assistant." It tests if the model remains consistent when presented with the target token by first providing an example of how it handles a nonsensical string.
\begin{quote}
\textbf{System Prompt:} \texttt{This helpful assistant explains what a phrase means, replying with identical responses to identical questions for reliability.} \\[1em]
\textbf{Conversation History:} \\
\textit{User:} \texttt{What does 'edslfkj' mean?} \\
\textit{Assistant:} \texttt{'edslfkj' has no specific meaning I am aware of.} \\[1em]
\textbf{Final User Prompt:} \texttt{What does '\{token\}' mean?} \\
\textbf{Assistant Prefill:} \texttt{`}
\end{quote}

\subsection{Analysis of the Verification Stage}
\label{sec:appendix_verification_analysis}

The GlitchMiner framework operates as a two-stage pipeline: 
\begin{enumerate}
    \item A gradient-guided search proposes candidate tokens that maximize output entropy.
    \item A robust verification stage confirms whether these candidates are true glitches.
\end{enumerate}

The verification stage is intentionally stringent, requiring a candidate to fail the repetition task across \textbf{all three} prompt templates (including those adapted from GlitchHunter and Magikarp) to be confirmed. A token that fails on only one or two templates does not meet this strict criterion.

Table~\ref{tab:filtered_candidates} summarizes the number of false-positive glitch tokens among the top 1000 candidates, which fail verification with template 1 but pass verification with either template 2 or template 3. This analysis demonstrates the necessity of employing cross-validation across multiple prompt templates.

\begin{table*}[h]
\centering
\begin{tabularx}{0.8\textwidth}{Xl}
\toprule
\textbf{Model} & \textbf{Candidates Filtered by Verification (Top 1000)} \\
\midrule
Qwen2.5-7B-Instruct & 0 \\
Qwen2-7B-Instruct & 0 \\
Llama-2-7b-chat-hf & 36 \\
Llama-3.1-8B-Instruct & 42 \\
Gemma-2-2b-it & 34 \\
Gemma-2-9b-it & 28 \\
Mistral-7B-Instruct-v0.3 & 51 \\
Mistral-Nemo-Instruct-2407 & 23 \\
Phi-3.5-mini-instruct & 76 \\
Phi-3-mini-128k-instruct & 49 \\
\bottomrule
\end{tabularx}
\caption{Count of high-entropy candidates from the top 1000 search results that were filtered out during the robust verification stage. These are tokens that did not fail on all three verification templates.}
\label{tab:filtered_candidates}
\end{table*}

\subsection{Rationale and Details of the Token Filter}
\label{sec:appendix_filter}

The core function of our Token Filter is to determine a token's practical relevance and, consequently, its real-world risk profile. It answers a fundamental question: \textbf{can a given token in the vocabulary be triggered by a user through standard text input?} If a token cannot be produced from text, it is excluded from our analysis. This is not part of the glitch detection logic itself, but a crucial prerequisite that allows GlitchMiner to focus its computational budget efficiently on tokens that represent genuine security and reliability concerns.

Our filter removes three main categories of such tokens:

\paragraph{SPECIAL Tokens} These are predefined functional symbols used for model structure or training (e.g., \texttt{[BOS]}, \texttt{</s>}). As they are not part of natural language inputs, they are straightforward to exclude.

\paragraph{Unreachable Tokens} These are tokens that exist in the vocabulary but can never be generated by the model's tokenizer from any text input, often existing as artifacts of the tokenizer's construction. We provide a concrete example from the \texttt{Llama-2-7B-chat-hf} model to illustrate this:
\begin{itemize}
    \item The token ID \texttt{71} corresponds to the bytecode \texttt{\textbackslash x44}, which the tokenizer decodes into the string ``D''.
    \item However, when we re-encode either the string ``D'' or the bytecode \texttt{\textbackslash x44}, the tokenizer consistently produces the token ID \texttt{360}, never 71.
\end{itemize}
This demonstrates that no text input a user can provide will ever result in token 71. Since it cannot be triggered in practice, testing it for "glitchiness" is unnecessary.

\paragraph{Undecodeable Tokens} This category includes tokens corresponding to byte sequences that cannot be decoded into a valid string, often due to violating encoding standards like UTF-8. For example, in the same Llama-2 tokenizer:
\begin{itemize}
    \item The token ID \texttt{255} corresponds to the byte \texttt{\textbackslash xFC}.
    \item This byte is an invalid start for a UTF-8 character sequence. Therefore, it cannot be decoded into coherent text and often appears as a generic replacement symbol.
\end{itemize}
Such tokens are excluded as they do not map to any meaningful user-generated text.

\section{Examples of Discovered Glitch Tokens}
\label{sec:appendix_examples}

Table~\ref{tab:glitch_examples} showcases several examples of glitch tokens discovered by GlitchMiner across different models, illustrating the variety of anomalous behaviors that can be triggered. The "Response" column displays each model's full output when given the primary verification prompt, \texttt{Please repeat the string: \{token\}}.


%% file: aaai2026.bib
@article{land2024fishing,
  title={Fishing for Magikarp: Automatically Detecting Under-trained Tokens in Large Language Models},
  author={Land, Sander and Bartolo, Max},
  journal={arXiv preprint arXiv:2405.05417},
  year={2024}
}

@article{LLM_code1,
  title={A Survey on Large Language Models for Code Generation},
  author={Jiang, Juyong and Wang, Fan and Shen, Jiasi and Kim, Sungju and Kim, Sunghun},
  journal={arXiv preprint arXiv:2406.00515},
  year={2024}
}

@article{LLM_code2,
  title={Evaluating large language models trained on code},
  author={Chen, Mark and Tworek, Jerry and Jun, Heewoo and Yuan, Qiming and Pinto, Henrique Ponde De Oliveira and Kaplan, Jared and Edwards, Harri and Burda, Yuri and Joseph, Nicholas and Brockman, Greg and others},
  journal={arXiv preprint arXiv:2107.03374},
  year={2021}
}

@article{LLM_code3,
  title={Codegen: An open large language model for code with multi-turn program synthesis},
  author={Nijkamp, Erik and Pang, Bo and Hayashi, Hiroaki and Tu, Lifu and Wang, Huan and Zhou, Yingbo and Savarese, Silvio and Xiong, Caiming},
  journal={arXiv preprint arXiv:2203.13474},
  year={2022}
}

@inproceedings{medical1,
  title={Llms accelerate annotation for medical information extraction},
  author={Goel, Akshay and Gueta, Almog and Gilon, Omry and Liu, Chang and Erell, Sofia and Nguyen, Lan Huong and Hao, Xiaohong and Jaber, Bolous and Reddy, Shashir and Kartha, Rupesh and others},
  booktitle={Machine Learning for Health (ML4H)},
  pages={82--100},
  year={2023},
  organization={PMLR}
}

@article{medical2,
  title={Augmenting black-box llms with medical textbooks for clinical question answering},
  author={Wang, Yubo and Ma, Xueguang and Chen, Wenhu},
  journal={arXiv preprint arXiv:2309.02233},
  year={2023}
}

@article{education1,
  title={Large language models for education: A survey and outlook},
  author={Wang, Shen and Xu, Tianlong and Li, Hang and Zhang, Chaoli and Liang, Joleen and Tang, Jiliang and Yu, Philip S and Wen, Qingsong},
  journal={arXiv preprint arXiv:2403.18105},
  year={2024}
}

@inproceedings{education2,
  title={Evaluating llm-generated worked examples in an introductory programming course},
  author={Jury, Breanna and Lorusso, Angela and Leinonen, Juho and Denny, Paul and Luxton-Reilly, Andrew},
  booktitle={Proceedings of the 26th Australasian Computing Education Conference},
  pages={77--86},
  year={2024}
}

@article{geiping2024coercing,
  title={Coercing LLMs to do and reveal (almost) anything},
  author={Geiping, Jonas and Stein, Alex and Shu, Manli and Saifullah, Khalid and Wen, Yuxin and Goldstein, Tom},
  journal={arXiv preprint arXiv:2402.14020},
  year={2024}
}

@misc{solidgoldmagikarp,
  title = {SolidGoldMagikarp (plus, prompt generation)},
  author = {{LessWrong Community}},
  year = {2023},
  howpublished = {\url{https://www.lesswrong.com/posts/aPeJE8bSo6rAFoLqg/so}},
  note = {Accessed: 2023-09-25}
}

@article{GlitchHunter,
  title={Glitch tokens in large language models: categorization taxonomy and effective detection},
  author={Li, Yuxi and Liu, Yi and Deng, Gelei and Zhang, Ying and Song, Wenjia and Shi, Ling and Wang, Kailong and Li, Yuekang and Liu, Yang and Wang, Haoyu},
  journal={Proceedings of the ACM on Software Engineering},
  volume={1},
  number={FSE},
  pages={2075--2097},
  year={2024},
  publisher={ACM New York, NY, USA}
}

@article{Magikarp,
  title={Fishing for Magikarp: Automatically Detecting Under-trained Tokens in Large Language Models},
  author={Land, Sander and Bartolo, Max},
  journal={arXiv preprint arXiv:2405.05417},
  year={2024}
}

@article{GlitchProber,
  title={GlitchProber: Advancing Effective Detection and Mitigation of Glitch Tokens in Large Language Models},
  author={Zhang, Zhibo and Bai, Wuxia and Li, Yuxi and Meng, Mark Huasong and Wang, Kailong and Shi, Ling and Li, Li and Wang, Jun and Wang, Haoyu},
  journal={arXiv preprint arXiv:2408.04905},
  year={2024}
}

@article{hotflip,
  title={Hotflip: White-box adversarial examples for text classification},
  author={Ebrahimi, Javid and Rao, Anyi and Lowd, Daniel and Dou, Dejing},
  journal={arXiv preprint arXiv:1712.06751},
  year={2017}
}

@article{autoprompt,
  title={Autoprompt: Eliciting knowledge from language models with automatically generated prompts},
  author={Shin, Taylor and Razeghi, Yasaman and Logan IV, Robert L and Wallace, Eric and Singh, Sameer},
  journal={arXiv preprint arXiv:2010.15980},
  year={2020}
}

@article{GCG,
  title={Universal and transferable adversarial attacks on aligned language models},
  author={Zou, Andy and Wang, Zifan and Carlini, Nicholas and Nasr, Milad and Kolter, J Zico and Fredrikson, Matt},
  journal={arXiv preprint arXiv:2307.15043},
  year={2023}
}

@article{shen2023anything,
  title={" do anything now": Characterizing and evaluating in-the-wild jailbreak prompts on large language models},
  author={Shen, Xinyue and Chen, Zeyuan and Backes, Michael and Shen, Yun and Zhang, Yang},
  journal={arXiv preprint arXiv:2308.03825},
  year={2023}
}

@article{traag2019louvain,
  title={From Louvain to Leiden: guaranteeing well-connected communities},
  author={Traag, Vincent A and Waltman, Ludo and Van Eck, Nees Jan},
  journal={Scientific reports},
  volume={9},
  number={1},
  pages={1--12},
  year={2019},
  publisher={Nature Publishing Group}
}

@misc{llama3_1,
  title = {Introducing Llama 3.1: Our most capable models to date},
  author = {Meta AI},
  year = {2024},
  howpublished = {\url{https://ai.meta.com/blog/meta-llama-3-1/}},
  note = {Accessed: 2024-10-18}
}

@article{llama2,
  title={Llama 2: Open foundation and fine-tuned chat models},
  author={Touvron, Hugo and Martin, Louis and Stone, Kevin and Albert, Peter and Almahairi, Amjad and Babaei, Yasmine and Bashlykov, Nikolay and Batra, Soumya and Bhargava, Prajjwal and Bhosale, Shruti and others},
  journal={arXiv preprint arXiv:2307.09288},
  year={2023}
}

@article{qwen2,
  title={Qwen2 technical report},
  author={Yang, An and Yang, Baosong and Hui, Binyuan and Zheng, Bo and Yu, Bowen and Zhou, Chang and Li, Chengpeng and Li, Chengyuan and Liu, Dayiheng and Huang, Fei and others},
  journal={arXiv preprint arXiv:2407.10671},
  year={2024}
}

@misc{qwen2_5,
  title = {Qwen 2.5: Advancing AI for Everyone},
  author = {Alibaba},
  year = {2024},
  howpublished = {\url{https://qwen2.org/qwen2-5/}},
  note = {Accessed: 2024-10-18}
}

@article{gemma2,
  title={Gemma 2: Improving open language models at a practical size},
  author={Team, Gemma and Riviere, Morgane and Pathak, Shreya and Sessa, Pier Giuseppe and Hardin, Cassidy and Bhupatiraju, Surya and Hussenot, L{\'e}onard and Mesnard, Thomas and Shahriari, Bobak and Ram{\'e}, Alexandre and others},
  journal={arXiv preprint arXiv:2408.00118},
  year={2024}
}

@article{phi3,
  title={Phi-3 technical report: A highly capable language model locally on your phone},
  author={Abdin, Marah and Jacobs, Sam Ade and Awan, Ammar Ahmad and Aneja, Jyoti and Awadallah, Ahmed and Awadalla, Hany and Bach, Nguyen and Bahree, Amit and Bakhtiari, Arash and Behl, Harkirat and others},
  journal={arXiv preprint arXiv:2404.14219},
  year={2024}
}

@article{mistral_7B,
  title={Mistral 7B},
  author={Jiang, Albert Q and Sablayrolles, Alexandre and Mensch, Arthur and Bamford, Chris and Chaplot, Devendra Singh and Casas, Diego de las and Bressand, Florian and Lengyel, Gianna and Lample, Guillaume and Saulnier, Lucile and others},
  journal={arXiv preprint arXiv:2310.06825},
  year={2023}
}

@misc{mistral2024nemo,
  title = {Mistral Nemo: Advancing the Capabilities of Large Language Models},
  author = {Mistral AI},
  year = {2024},
  howpublished = {\url{https://mistral.ai/news/mistral-nemo/}},
  note = {Accessed: 2024-10-18}
}

@article{PEZ,
  title={Hard prompts made easy: Gradient-based discrete optimization for prompt tuning and discovery},
  author={Wen, Yuxin and Jain, Neel and Kirchenbauer, John and Goldblum, Micah and Geiping, Jonas and Goldstein, Tom},
  journal={Advances in Neural Information Processing Systems},
  volume={36},
  year={2024}
}
